%% file: main_arxiv.tex
\documentclass[letterpaper]{article} 
\usepackage[draft]{aaai25}  
\usepackage{times}  
\usepackage{helvet}  
\usepackage{courier}  
\usepackage[hyphens]{url}  
\usepackage{graphicx} 
\usepackage{natbib}  
\usepackage{caption} 
\usepackage{algorithm}
\usepackage{algorithmic}

\usepackage{todonotes}
\usepackage{booktabs}
\usepackage{multirow}
\usepackage{amsmath}
\usepackage{amsfonts}
\usepackage{amssymb}
\usepackage{colortbl} 
\usepackage{xcolor} 
\definecolor{customteal}{rgb}{0.2, 0.6, 0.9} 
\definecolor{customlightblue}{rgb}{0.82, 0.94, 1.0} 

\newcommand{\tabglm}{\mbox{TabGLM}}
\usepackage{newfloat}
\usepackage{listings}
\DeclareCaptionStyle{ruled}{labelfont=normalfont,labelsep=colon,strut=off} 
\floatstyle{ruled}
\newfloat{listing}{tb}{lst}{}
\floatname{listing}{Listing}
\title{\tabglm: Tabular Graph Language Model for Learning \\ Transferable Representations Through Multi-Modal Consistency Minimization}
\author{
    Anay Majee\textsuperscript{\rm 1, \rm 2}\equalcontrib \thanks{Work done as an intern at Fujitsu.},
    Maria Xenochristou\textsuperscript{\rm 1}\equalcontrib,
    Wei-Peng Chen\textsuperscript{\rm 1}
}
\affiliations{
    \textsuperscript{\rm 1}Fujitsu Research of America \\
    \textsuperscript{\rm 2}The University of Texas at Dallas \\
    anay.majee@utdallas.edu, mxenochristou@fujitsu.com, wchen@fujitsu.com
}

\begin{document}

\maketitle

\begin{abstract}
\input{sections/01-abstract}
\end{abstract}

%

\section{Introduction}
\input{sections/02-introduction}

\section{Related Work}
\label{sec:related_work}
\input{sections/03-related_work}

\section{Method}
\label{sec:method}
\input{sections/04-method}

\section{Experiments}
\label{sec:experiments}
\input{sections/05-experiment}

\section{Conclusion}
\input{sections/06-conclusion}

\section*{Acknowledgments}
We gratefully thank anonymous reviewers for their valuable comments. 
We would also like to extend our gratitude to the members of the AI Lab at Fujitsu Research of America (FRA) for their valuable feedback and constructive criticism during the development of the project. 
Additionally, Anay Majee would like to thank the leadership at FRA for providing the opportunity to intern at FRA.

\bibliography{references}

\newpage
\appendix
\section*{Appendix}
\input{sections/07-appendix}

\end{document}

%% file: sections/01-abstract.tex
Handling heterogeneous data in tabular datasets poses a significant challenge for deep learning models. While attention-based architectures and self-supervised learning have achieved notable success, their application to tabular data remains less effective over linear and tree based models.
Although several breakthroughs have been achieved by models which transform tables into uni-modal transformations like image, language and graph, these models often underperform in the presence of feature heterogeneity.
To address this gap, we introduce \textbf{\tabglm} (\textbf{Tab}ular \textbf{G}raph \textbf{L}anguage \textbf{M}odel), a novel multi-modal architecture designed to model both structural and semantic information from a table. 
\tabglm\ transforms each row of a table into a fully connected graph and serialized text, which are then encoded using a graph neural network (GNN) and a text encoder, respectively. By aligning these representations through a joint, multi-modal, self-supervised learning objective, \tabglm\ leverages complementary information from both modalities, thereby enhancing feature learning.
\tabglm's flexible graph-text pipeline efficiently processes heterogeneous datasets with significantly fewer parameters over existing Deep Learning approaches. Evaluations across 25 benchmark datasets demonstrate substantial performance gains, with \tabglm\ achieving an average AUC-ROC improvement of up to 5.56\% over State-of-the-Art (SoTA) tabular learning methods.\looseness-1

%% file: sections/02-introduction.tex
Real-world applications ranging from predicting sales in e-commerce to diagnosing diseases in healthcare rely on tabular data. These datasets are oftentimes a mix of numerical, categorical, and text values, presenting a unique challenge for machine learning models. 
Traditional approaches~\cite{breiman2001random, chen2016xgboost, prokhorenkova2018catboost} as well as some early Deep Learning (DL) models~\cite{yoon2020vime, arik2021tabnet, gorishniy2021revisiting, hollmann2022tabpfn} convert textual data into numerical encodings modeling only structural features from an input table, leading to loss of semantic information.
Recent trends in tabular DL indicate an increase in approaches attempting modality switch from tabular to image~\cite{DeepInsight, wang2019supertml}, text~\cite{tabllm, arik2021tabnet}, or graph~\cite{ignnet, tabgnn}, modeling \textit{either} semantic or structural relationships. These transformations aim to exploit the strengths of established models in vision, language, and graph domains to enhance the representation learning of tabular data. 
Unfortunately, modeling a single type of relationship through uni-modal transformation limits the ability of DL models in this domain to perform well on heterogeneous datasets. In addition, DL models are often prone to overfitting, especially on datasets with high dimensionality or limited samples. Thus, such models frequently struggle to outperform simple linear and tree based models. This discrepancy highlights a fundamental challenge: \textit{effectively integrating the diverse types of information within tabular data}, while preserving the rich semantic and structural nuances.

\input{figures/title_figure}

We bridge this gap by introducing \textbf{\tabglm} (\textbf{Tab}ular \textbf{G}raph \textbf{L}anguage \textbf{M}odel), a novel multi-modal architecture designed to effectively capture \textbf{both structural and semantic information in tabular data}. This is achieved by transforming each row of a tabular dataset into a graph and serialized text, and encoding it using a graph neural network (GNN) and a pretrained text encoder, respectively (Figure \ref{fig:title_figure}). 
Transforming a record into a graph encodes relationships between columns, thus modeling structure, while transforming it into text embeddings captures semantic information.
The joint semi-supervised learning strategy in \tabglm, namely \textsc{MuCosa} (detailed in Section \ref{sec:joint_learner}), aligns the learned representations from both the graph and text encoders while adapting to downstream tasks.
This alignment enhances the quality of the learnt representations by leveraging complementary information from both modalities, while acting as a regularization strategy to prevent overfitting.

To the best of our knowledge, we are the \textit{first to introduce a multi-modal learning framework for tabular data}, with the following principal contributions -
\begin{itemize}
    \item We introduce a \textbf{multi-modal method} that transforms each row of a tabular dataset into a graph and serialized text, capturing both structural and semantic features.\looseness-1
    \item Our joint loss (\textsc{MuCosa}) assists with information fusion from the 2 modalities, while acting as a regularization mechanism to mitigate overfitting.
    \item \tabglm's~targeted use of frozen and trainable components achieves a \textbf{significantly lower (by over 80\%) parameter count} compared to State-of-the-Art (SoTA) uni-modal DL approaches.\looseness-1
    \item Extensive experiments and ablation studies validate the effectiveness of \tabglm\ and its key components. We \textbf{demonstrate an absolute improvement in AUC-ROC scores up to 5.56\%, compared to SoTA models across 25 benchmark datasets} detailed in Section \ref{sec:experiments} of the main paper.\looseness-1
\end{itemize}

%% file: figures/title_figure.tex
\begin{figure}[t]
        \centering
        \includegraphics[width=\columnwidth]{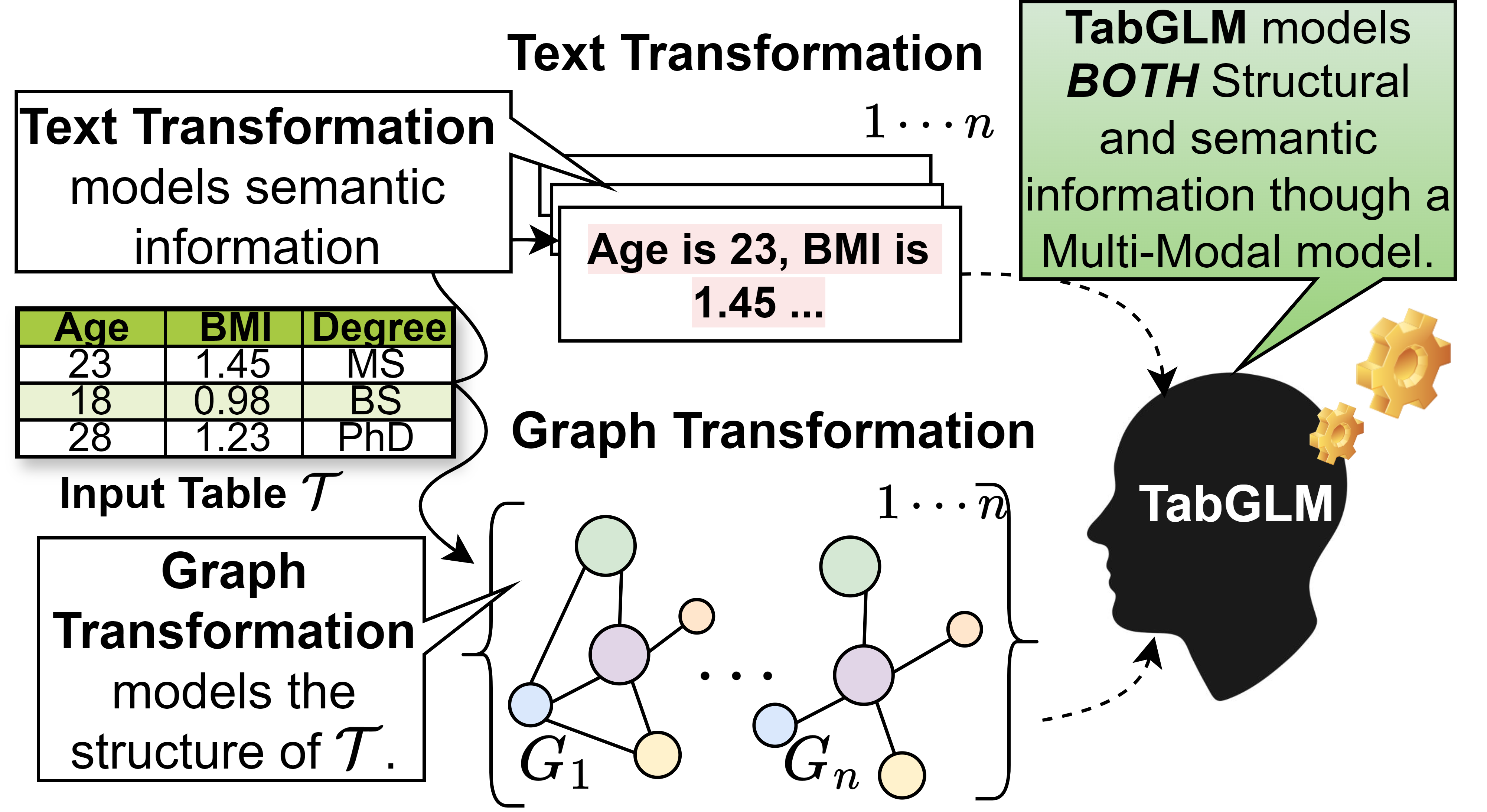}
        \caption{\textbf{Semi-Supervised Multi-Modal Tabular Deep Learning in \tabglm.} We propose a joint graph-language method that can effectively learn from heterogeneous, real-world tabular datasets by integrating structural and semantic information.\looseness-1}
        \label{fig:title_figure}
\end{figure}

%% file: sections/03-related_work.tex
\noindent \textbf{Traditional Tabular Machine Learning} 
Historically, the realm of tabular data modeling over the past decade has been largely dominated by conventional machine learning methods~\cite{shwartz2022tabular}. 
Models such as Gradient Boosting~\cite{bentejac2021comparative}, ExtraTrees~\cite{geurts2006extremely}, and Random Forests~\cite{breiman2001random} have been pivotal in learning intricate data patterns and enhancing robustness against overfitting. 
Notable techniques like XGBoost \cite{chen2016xgboost} and LightGBM \cite{ke2017lightgbm} stand out for their efficiency, optimization techniques, and scalability, making them go-to options in various applications. 
Logistic regression \cite{hosmer2013applied} has been particularly applied to binary classification tasks due to its simplicity and interpretability.
Specialized algorithms like CatBoost~\cite{prokhorenkova2018catboost}, designed to handle categorical features seamlessly, have gained prominence. 
This diverse set of models contributes to a versatile toolbox, addressing the intricacies of tabular data modeling with distinct strengths and adaptability \cite{grinsztajn2022tree}. These traditional models provide a solid foundation for tabular data analysis, balancing interpretability, efficiency, and performance essential for real-world applications.
Despite their effectiveness, these models are often limited by their reliance on handcrafted feature engineering and their inability to leverage the representation learning capabilities inherent in deep learning models.\looseness-1

\input{figures/overview}

\noindent \textbf{Transformers for tabular data} 
Following the popularity of Transformer architectures in vision and language, several methods~\cite{hollmann2022tabpfn, arik2021tabnet, zhu2023xtab} have adapted transformers for learning from tabular datasets. 
For instance, FT-Transformer~\cite{gorishniy2021revisiting} showed superior performance in tabular classification and regression tasks by separating numerical and categorical features. 
Additionally, Saint~\cite{somepalli2021saint} introduced row-wise attention, capturing inter-sample interactions, Fastformer~\cite{wu2021fastformer} suggested the use of additive attention which is lightweight with linear complexity, while TransTab~\cite{wang2022transtab} incorporated transfer learning in tabular tasks, all using transformers as backbones. Recent advancements have specifically tailored the transformer architecture to address challenges in data imputation and cross-table learning, incorporating modifications to the attention mechanism and embedding layers~\cite{badaro2023transformers}.\looseness-1

\noindent \textbf{Self-supervised pretraining} 
Furthermore, the emergence of self-supervised pretraining in the tabular domains has paved the way for novel approaches to feature extraction and representation learning, reducing the reliance on labeled data \cite{liu2021self}. Specifically, drawing inspiration from the success of pretraining in vision and language, previous studies have delved into tabular self-supervised learning \cite{yoon2020vime, ucar2021subtab, somepalli2021saint, bahri2021scarf, majmundar2022met, rubachev2022revisiting, wang2022transtab}. 
Authors in~\cite{yoon2020vime, ucar2021subtab} introduced an auto-encoder framework with a pretext task focused on reconstructing missing elements in a table while \cite{bahri2021scarf} utilized contrastive learning~\cite{chen2020simple} as pretraining objective for improving generalizability of trained architectures in tabular tasks. 
Additionally, \cite{rubachev2022revisiting, wang2022transtab} created a target-aware objective by incorporating label columns of tabular tasks in pretraining. 
Although these innovations have largely improved performance over traditional machine learning approaches, these models have been shown to particularly underperform in the presence of heterogeneous feature columns~\cite{tabllm}.\looseness-1

\noindent \textbf{Modality switch for Tabular Deep Learning} Recent research has explored the conversion of tabular data into orthogonal modalities, such as text, image, and graph. 
TabLLM \cite{tabllm} converted tabular data to text for few-shot classification using large language models. Although it can suffer from context loss and inefficiency when handling high-dimensional data, TabLLM successfully captures the semantic information encapsulated within columns in a table.
SuperTML \cite{wang2019supertml} introduced a method to transform tabular data into a super ensemble of image-based data points, enabling the use of convolutional neural networks for tabular tasks. DeepInsight \cite{DeepInsight} proposed projecting tabular data into an image space using t-SNE, enabling the application of image classification models to tabular data. Even though this technique effectively captures underlying feature correlations, the reliance on a single-image representation and t-SNE's specific distance metric limits its ability to capture diverse and multi-faceted relationships inherent in complex tabular datasets. 
Table2Graph~\cite{huang22table2graph} transforms tabular data into a unified weighted graph and IGNNet \cite{ignnet} transforms each record into a fully-connected graph, allowing the application of graph neural networks (GNNs) for tabular data learning. 
Additionally, GCondNet~\cite{margeloiu2023gcondnet} transforms each column into a graph while CARTE~\cite{kim2024carte} mines entities in tables to learn from entity-centric graphs.
Furthermore, models like Graph foundation models \cite{galkin2024towards, zhang2024gnn} and \cite{sun2023gpt} highlight the efficacy of GNNs in capturing relational structures within tabular data.  
HyTrel \cite{chen2023hytrel} enhances tabular data representation by integrating hypergraph structures, which can capture high-order relationships among features, but the complexity of hypergraph construction and the increased computational cost are significant challenges.
Despite their innovative approach, these methods often face scalability issues with large datasets and are sensitive to the graph construction method. Additionally, even though graphs can capture the structural relationships among features in a table, they cannot capture the semantic information of the categorical and text columns, as well as the column headers. This information can provide valuable insight, which is especially valuable when learning from small datasets. \looseness-1

\noindent \textbf{Multi-Modal Learning}
Multi-modal learning integrates data from multiple sources, such as text, image, video, and audio to enhance machine learning models' performance. A pivotal model in this domain is CLIP \cite{radford2021learning}, which aligns text and image representations using contrastive learning, enabling effective zero-shot learning and image-text retrieval. Other significant advancements include~\cite{hegde2023clip3d}, which adapts CLIP to 3D recognition tasks through prompt tuning for language grounding, as well as \cite{Chen_2023}, which introduces cross-modal knowledge distillation, and \cite{ramesh2021zero}, which introduces zero-shot text-to-image generation.\looseness-1

An important lesson from existing literature is that multi-modal models are capable of generalizing to downstream tasks by capturing complementary information from multiple modalities. For instance, they extract complex spatial patterns from images, semantic meaning from text, and structural relationships from graphs.  
We capitalize on this property to design a multi-modal model for tabular machine learning that combines the richness of graph and text modalities into a unified embedding space to improve performance on downstream ML tasks.
To the best of our knowledge, we are the first to introduce multi-modal learning for tabular datasets using a single table as input across several classification based downstream tasks.\looseness-1

%% file: figures/overview.tex
\begin{figure*}[t]
        \centering
        \includegraphics[width=\textwidth]{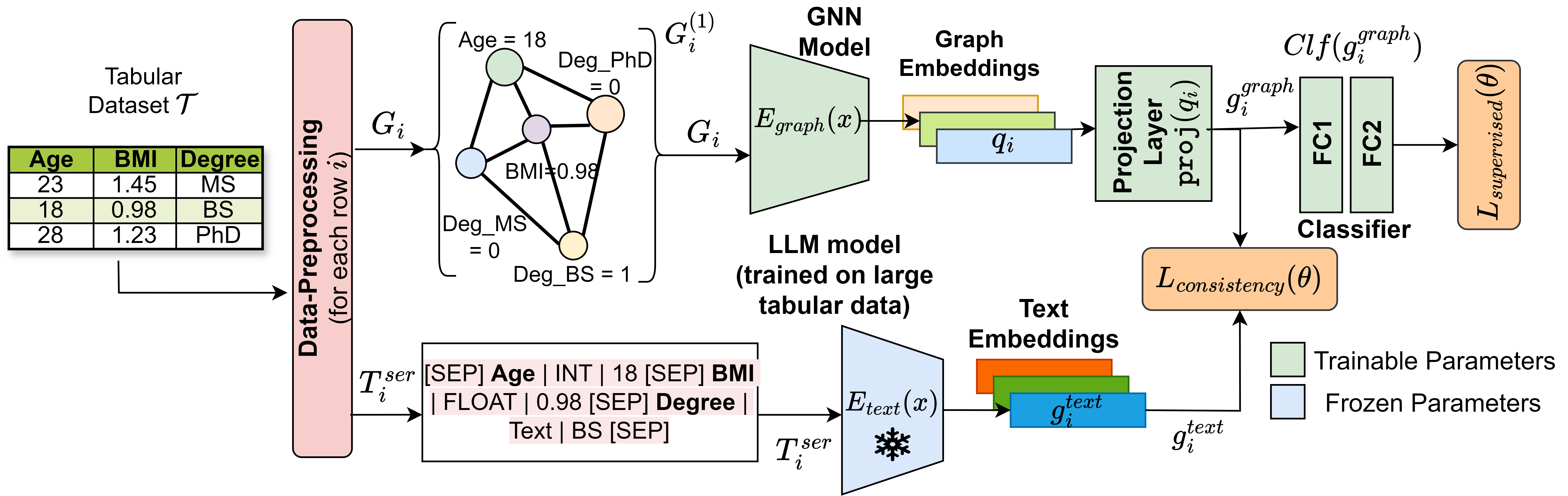}
        \caption{\textbf{Overview of our \tabglm\ framework.} \tabglm\ introduces Multi-modal Graph-Language Modeling to enable tabular learning on datasets with heterogeneous data types. Our method leverages graph and language embeddings, consistency regularization, and supervised learning to effectively adapt to diverse real-world downstream tasks.\looseness-1}
        \label{fig:tabglm_overview}
\end{figure*}

%% file: sections/04-method.tex
Datasets in real-world are oftentimes heterogeneous consisting of both numerical and textual features. To this end, we introduce Tabular Graph Language Model (\tabglm) depicted in Figure \ref{fig:tabglm_overview}, which tackles the aforementioned challenge by preserving both structural and semantic features enumerated in tabular datasets.

\subsection{Problem Definition}
\label{sec:problem_definition}
Given a tabular dataset $\mathcal{T} \in \mathbb{R}^{n \times m}$ represented as a matrix of $n$ records each with $m$ feature columns and a label $y_i$, where $i \in |\mathcal{T}|$, we are tasked to predict the probability of a newly introduced record $x$ in the test dataset to be one among the target classes $y_i$.
To achieve this goal we train a feature representation learner $h(\mathcal{T}_i; \theta)$, which learns feature representations $g_i$ from samples (rows) $\mathcal{T}_i$ in a training dataset $\mathcal{T}_{train}$, where $i \in [1, n]$ and total model parameters $\theta$. The learned representations $g_i, \forall i \in [1,n]$ are then passed to a predictor $Clf(g_i)$ for downstream classification tasks. 
The performance of the model $h(x_i, \theta)$ on unseen records in $\mathcal{T}_{test}$ largely depends on the quality of learned representations $g$.
Particularly, in this paper we tackle the challenges associated with tables $\mathcal{T}$ containing heterogeneous column types (both numeric and textual) through a multi-modal architecture as discussed in Section \ref{sec:tabglm}.

\subsection{\tabglm: Tabular Graph Language Model}
\label{sec:tabglm}
\tabglm\ introduces a multi-modal architecture as shown in Figure \ref{fig:tabglm_overview}, which encodes each record in an input table into a graph (learning structural features) and serialized text (learning semantic features).
As highlighted in Section \ref{sec:related_work}, recent approaches employ uni-modal transformations, encoding either structural or semantic information. 
A uni-modal model, therefore, lacks the advantages provided by auxiliary modalities (promoting learning of only specific types of features). The multi-modal architecture of \tabglm\ addresses this gap and demonstrates improvements in downstream tasks by combining two complimentary modalities in a single unified architecture. 

We simultaneously transform each record $\mathcal{T}_i \in \mathcal{T}$ into a fully-connected graph $G_i$ and natural language (serialized text) $T_i^{ser}$. 
\tabglm\ then encodes $G_i$ and $T_i^{ser}$ using a graph encoder $E_{graph}$ and a text encoder $E_{text}$, producing feature vectors $g_i^{graph}$ and $g_i^{text}$ respectively.
Finally, we combine the encoded feature vectors, $g_i^{graph}$ and $g_i^{text}$ using a Multi-Modal Consistency Learner (MuCosa) that minimizes the feature separation between complimentary modalities (unsupervised) while adapting to downstream tasks (supervised).
We detail the aforementioned components in our \tabglm\ architecture below, which can be decomposed into the text pipeline and the graph pipeline.

\input{figures/data_processing}

\noindent \textbf{Text Pipeline } The text encoder ($E_{text}$) encodes each record $\mathcal{T}_i \in\mathcal{T}$ into an embedding $g_i^{text}$ with the goal of preserving the semantic information in the cell values. 
We achieve this by first transforming each row in $\mathcal{T}$ to serialized natural text $T^{ser}$ as shown in Figure \ref{fig:data_preprocessing}. 
Drawing inspiration from the authors in \citet{tabllm}, we adopt the simple \texttt{text} serialization which is inexpensive while being representative. An example of this is also depicted in Figure \ref{fig:data_preprocessing} where each row in the table in represented as a templated~\cite{tabllm, tapas} text (serialization). 
The serialized output is tokenized to convert the natural language input (each row in the input table) into tensors $T^{ser}$ before passing to the text encoder $E_{text}$. The text encoder produces the text embedding $g_i^{text}$ as shown in Equation \ref{eq:text_encoding}, where $g_i^{text} \in \mathbb{R}^d$.
Following the recent success of LLMs in tabular question answering~\cite{tapex, tapas}, \tabglm\ adopts the best performing pretrained text encoders in TAPAS~\cite{tapas} and TAPEX~\cite{tapex}, trained on a large number of records. 
The choice of the text encoder presents a trade-off between performance and computational complexity, which is elucidated through the ablation study in Section \ref{sec:ablations}.
The parameters $\theta_{text}$ of the text encoder $E_{text}$ are kept frozen during the training process and used to produce an instance-level (row) embedding \textbf{$g_i^{text}$, encoding context aware features from each record}.
\begin{align}
\begin{split}
g_i^{text} = E_{text}(\texttt{tokenize}(\mathcal{T}_i^{ser}); \theta_{text})
\end{split}
\label{eq:text_encoding}
\end{align}

\noindent \textbf{Graph Pipeline } The Graph Encoder ($E_{graph}$) takes as input a fully connected graph $G(v, e)$ to learn embeddings $g_i^{graph}$ corresponding to each row $\mathcal{T}_i \in \mathcal{T}$. The goal of $E_{graph}$ is to encode the latent structural relationships between columns in the underlying table $\mathcal{T}$.
Following the authors in \citet{ignnet}, we first transform each record $\mathcal{T}_i \in \mathcal{T}$ into a graph representation $G_i(v, e)$ where each node in $v$ encodes the value associated with a feature column in $\mathcal{T}$ and each edge in $e$ represents the relationship between feature columns (as edge weight). 
During implementation, a set of edge weights $W$ re-weights each edge in $G_i$ while an adjacency matrix $A$ encodes the structure of $G_i$.
A known limitation of SoTA tabular graph learning approaches~\cite{ignnet, huang22table2graph} is the ability to encode categorical features. As depicted in Figure \ref{fig:data_preprocessing}, \tabglm\ converts columns with categorical features to numerical encodings as a preprocessing step before passing them as input to $E_{graph}$ which is parameterized by $\theta_{graph}$ as shown in Equation \ref{eq:graph_encoding}.\looseness-1
\begin{align}
\begin{split}
g_i^{graph} = \texttt{proj}( E_{graph}( G_i(v, e); \theta_{graph} ) )    
\end{split}
\label{eq:graph_encoding}
\end{align}

\tabglm\ learns a latent representation $q_i$ for each row in $\mathcal{T}$ by adopting the popular Graph Neural Network (GNN) in \citet{xu2019gnn, ignnet} which employs a sequence of \textit{message passing}~\cite{xu2019gnn} layers in its architecture to learn node level features. These node level features are further aggregated using \textit{read-out layers} to learn an embedding for the input graph $G_i$. 
Several iterations of message passing during model training allows \textbf{$E_{graph}$ to learn the structure of the table $\mathcal{T}$ by modeling the relationships between features columns} (represented as nodes $v$).
The output latent representation $q_i$ is projected to a lower dimensional space using a projection layer \texttt{proj} to produce graph embeddings $g_i^{graph} = \texttt{proj}(q_i)$ as depicted in Figure \ref{fig:tabglm_overview} and expressed in Equation \ref{eq:graph_encoding}. This layer projects the graph embedding in the same dimensional space as the text embedding.\looseness-1

During the \textit{training phase $E_{graph}$ is trained from scratch in an end to end fashion while $E_{text}$ remains frozen}, with the total parameter count of the feature extractor $h$ (composed of $E_{graph}$ and $E_{text}$) being $\theta = \theta_{graph} + \theta_{text}$. This design choice is based on the assumption that the text encoder in TAPAS / TAPEX has learnt generalizable representations from a large volume of records (26.9 billion in \citet{tapas})  it was pretrained on.ng \textit{inference, \tabglm\ omits the forward pass through the text encoder $E_{text}$}, relying solely on the embeddings learnt from $E_{graph}$, \textbf{significantly boosting inference speeds}.
Further, we show through ablation experiments in Section \ref{sec:ablations} that the proposed \textbf{\tabglm\ architecture uses only 336M parameters which is over 80\% lower than SoTA approach TabLLM}~\cite{tabllm}.\looseness-1

\subsection{\textsc{MuCosa}: \textbf{Mu}lti-Modal \textbf{Co}n\textbf{s}istency Le\textbf{a}rner}
\label{sec:joint_learner}
The training of \tabglm\ proceeds in a single stage with a joint sem-supervised learning approach, \textsc{MuCosa}.
As discussed in Section \ref{sec:tabglm}, the representations learnt from both $E_{graph}$ and $E_{text}$ encode orthogonal concepts with the former encoding structure and the latter encoding semantic information.
To combine the learnings from both $g_i^{graph}$ and $g_i^{text}$ we minimize the consistency between the two modalities through a consistency loss, $L_{\text{consistency}}$ as shown in Equation \ref{eq:consistency_loss}. $L_{\text{consistency}}$ aligns the text embeddings $g_i^{text}$ with the graph embeddings $g_i^{graph}$ corresponding to each row in $T_{train}$ and vice versa, in a label free fashion.\looseness-1
\begin{align}
\begin{split}
L_{\text{consistency}} = -\frac{1}{2n} \sum_{i=1}^{n} \Biggl[ \log \frac{\exp\left(\frac{\hat{g}_i^{\text{text}} \cdot (\overline{\hat{g}_i^{\text{graph}}})^T}{\tau}\right)}{\sum_{j=1}^{n} \exp\left(\frac{\hat{g}_i^{\text{text}} \cdot (\overline{\hat{g}_j^{\text{graph}}})^T}{\tau}\right)} \\
+ \log \frac{\exp\left(\frac{\hat{g}_i^{\text{graph}} \cdot (\overline{\hat{g}_i^{\text{text}}})^T}{\tau}\right)}{\sum_{j=1}^{n} \exp\left(\frac{\hat{g}_i^{\text{graph}} \cdot (\overline{\hat{g}_j^{\text{text}}})^T}{\tau}\right)} \Biggr] \\
\end{split}
\label{eq:consistency_loss}
\end{align}

Here, $\hat{g}_i^{\text{text}} = \frac{g_i^{\text{text}}}{\|g_i^{\text{text}}\|_2}$ and $\hat{g}_i^{\text{graph}} = \frac{g_i^{\text{graph}}}{\|g_i^{\text{graph}}\|_2}$ represents the normalized form of the graph and text embeddings, $\tau$ (set to 0.1 following \citet{chen2020simple}) denotes the temperature term and $\overline{\hat{g}_i^{\text{text}}}$, $\overline{\hat{g}_i^{\text{graph}}}$ indicates explicitly that gradients are not propagated for those terms.
Additionally, we minimize a supervised loss $L_{\text{supervised}}$ between the ground truth $y_i$ and the predicted logits from a classifier head $\hat{y}_i = Clf(g_i^{graph})$ as shown in Figure \ref{fig:tabglm_overview}. Note, that the classifier head consumes only the graph embeddings to mimic the inference setting.
The supervised loss can be represented as Equation \ref{eq:supervised}.\looseness-1

\begin{align}
    L_{\text{supervised}} = \frac{1}{n} \sum_{i = 1}^{n} H(y_i, \hat{y}_i)
\label{eq:supervised}
\end{align}

Finally, \tabglm\ introduces joint objective, \textsc{MuCosa} ($L$) as shown in Equation \ref{eq:joint_loss} which combines both $L_{\text{supervised}}$ and $L_{\text{consistency}}$ as a weighted sum with the hyper-parameter $\lambda$ controlling the contribution of each component to the total loss $L$.\looseness-1
\begin{align}
L = (1 - \lambda) L_{\text{supervised}} + \lambda L_{\text{consistency}}
\label{eq:joint_loss}
\end{align}

%% file: figures/data_processing.tex
\begin{figure}[t]
        \centering
        \includegraphics[width=0.90\columnwidth]{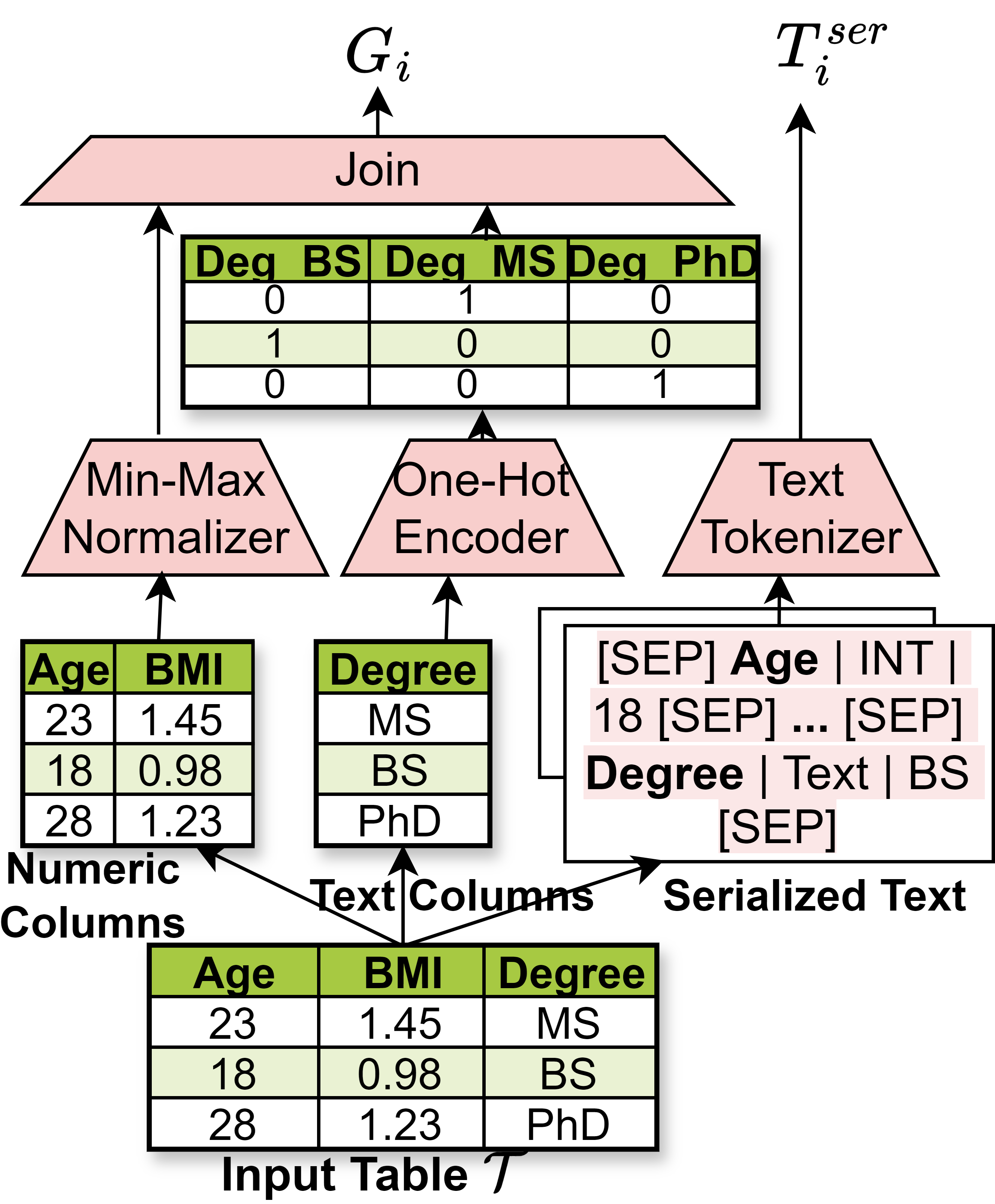}
        \caption{\textbf{Multi-Modal Representation of Tables in \tabglm,} depicting the text serialization and graph pre-processing pipelines.}
        \label{fig:data_preprocessing}
\end{figure}

%% file: sections/05-experiment.tex
We conduct our experiments on a wide variety of tabular datasets (refer Section \ref{sec:datasets}) with varying levels of data heterogeneity across a variety of downstream classification tasks (refer Section \ref{sec:tabglm}).

\input{tables/01-benchmark_Sota_methods}

\subsection{Datasets}
\label{sec:datasets}
To demonstrate the effectiveness of \tabglm\ in the presence of heterogeneous feature columns, as discussed in Section \ref{sec:tabglm}, we conduct our experiments on 25 datasets encompassing both binary and multi-class classification tasks, curated from popular papers TabLLM~\cite{tabllm}, TabPFN~\cite{hollmann2022tabpfn}, and large scale datasets in OpenML~\cite{openml2017}.
Following the principal goal of \tabglm, we consider heterogeneous datasets that encapsulate both numerical and textual columns like \textbf{Bank} ($\sim$45k records with 7 numerical and 9 categorical columns), \textbf{Creditg} (1k rows with 7 numerical and 13 categorical columns), \textbf{Heart} (918 rows with 6 numerical and 5 categorical columns) and \textbf{Income} ($\sim$48k rows with 4 numerical and 8 categorical columns), as shown in TabLLM. In addition, we use 12 datasets from OpenML, containing at least 1 numerical and 1 categorical column, including \textbf{balance-scale} (5 numerical and 1 categorical), \textbf{tic-tac-toe} (10 numerical and 10 categorical), \textbf{dress-sales} (13 numerical and 12 categorical) etc with more details in appendix. 
We also include datasets containing only numerical columns like \textbf{blood} (4 numerical columns), \textbf{calhousing} (8 numerical columns), \textbf{coil2000} (86 numerical columns) etc. alongside datasets containing only categorical columns like \textbf{car}, from both OpenML and TabPFN. 
We adopted datasets of varying sizes, with number of rows ranging from 500 (in \textbf{dress-sales}) to 45,211 (in \textbf{bank}) to demonstrate the applicability of our method to real-world large tabular datasets.
Note that the multi-modal architecture in \tabglm\ involves a LLM encoder~\cite{tapas, tapex} that is limited by the number of input tokens, which is 512 (from TAPAS) in our case.
More details on each dataset experimented upon in Table \ref{tab:sota_perf_contrast} is discussed in the supplementary material.\looseness-1

\subsection{Experimental Setup}
We conduct our experiments on datasets discussed in Section \ref{sec:datasets} and report the average performance (AUC-ROC scores) of each model across the same 5 random seeds (kept constant across datasets) in Section \ref{sec:results}. 
For all numerical and heterogeneous datasets, numerical columns are normalized using min-max\footnote{Scikit learn package: \url{https://scikit-learn.org/stable/modules/generated/sklearn.preprocessing.MinMaxScaler.html}} normalization while the categorical (text) columns are converted into One-Hot encodings (refer ablation in supplementary material) to create a numeric dataset for graph transformation.
For the text transformation, each record in the table is converted to serialized text following the tokenizer in TAPAS~\cite{tapas}. We chose TAPAS based on ablation experiments on the choice of LLMs in Section \ref{sec:ablations}. 
For datasets that contain only categorical columns, our \tabglm\ method uses only the text pipeline, utilizing only the semantic information present in such datasets.
Models for all datasets are trained on a fixed set of hyperparameters with an initial learning rate of $1e^{-4}$, batch size of 256 and weighting the consistency loss at 20\% ($\lambda = 0.2$). 
All experiments are conducted on 4 NVIDIA V100 GPUs with additional details on the experiment setup in the Appendix and code released at \url{https://github.com/amajee11us/TabGLM}.\looseness-1

\input{tables/02-benchmark_dl_methods}

\subsection{Results}
\label{sec:results}
At first, we compare the performance of \tabglm\ with traditional linear and tree based Machine Learning models like CatBoost~\cite{prokhorenkova2018catboost}, XGBoost~\cite{chen2016xgboost}, Gradient Boosting (GB) \cite{ke2017lightgbm}, Random Forest (RF) \cite{breiman2001random} and Logistic Regression (LR).
Our results in Table \ref{tab:sota_perf_contrast} show that \tabglm\ demonstrates significant increase in AUROC of 4.77\% over LR, 2.51\% over RF etc. outperforming such techniques across 25 downstream tabular classification tasks.
However, for simple datasets with lower number of feature columns like \textbf{kr-vs-kp}, \textbf{pc3} etc., tree based models (CatBoost) continue to show dominance in performance.\looseness-1

Secondly, we compare the performance of \tabglm\ with SoTA tabular DL models like FT-Transformer~\cite{gorishniy2021revisiting}, TabTransformer~\cite{huang2020tabtransformer} and NODE~\cite{popov2019neural}. \tabglm\ consistently outperforms tabular DL models like FT-Transformer by 5.56\%, TabTransformer by 3.64\% and NODE by 1.26\% respectively.
Finally, we compare the performance of \tabglm\ against SoTA uni-modal DL architectures like IGNNet (table-to-graph) and TabLLM (table-to-text) on 9 datasets in the benchmark introduced in TabLLM~\citet{tabllm}. We observe that \tabglm\ outperforms TabLLM by 1.35\% and IGNNet by 7.96\% respectively on the benchmark datasets in \cite{tabllm}, summarized in Table \ref{tab:dl_model_benchmark}.
The above results indicate a strong generalization of the proposed \tabglm\ architecture to a variety of downstream tasks, establishing \tabglm\ as a strong choice for Tabular Deep Learning under feature heterogeneity.\looseness-1

\subsection{Ablation Study}
\label{sec:ablations}

\input{tables/04-ablation_choice_of_llm}

\noindent \textbf{Multi-Modal vs. Uni-Modal training:}
The core contribution of \tabglm\ lies in its multi-modal architecture for tabular representation learning. To evaluate its components, we decompose it into two uni-modal architectures: \textit{Graph only} (using only the graph encoder $E_{\text{graph}}$) and \textit{Text only} (using only the text encoder $E_{\text{text}}$), based on the choice of the feature extractor during both training and inference. 
Their performance is compared against the complete multi-modal \tabglm\ training recipe, with results summarized in Table \ref{tab:tabglm_components}. 
The \textit{Graph only} pipeline employs the GNN from \cite{ignnet}, while the \textit{Text only} pipeline uses the BART-based TAPAS~\cite{tapas} encoder. 
Both pipelines use the same classifier head (Section \ref{sec:tabglm}) for downstream tasks. In \textbf{Text only}, the encoder is frozen, and only the classifier head is trained, whereas in \textit{Graph only}, both the encoder and classifier head are trained, to ensure fair comparison with \tabglm, where the text encoder remains frozen during training. 
Experiments on three representative datasets—\textbf{pc3} (numerical), \textbf{bank} (balanced numerical and categorical), and \textbf{creditg} (categorical-heavy)—show that \tabglm's multi-modal design consistently outperforms its uni-modal variants, underscoring the value of modality fusion for learning from heterogeneous tables.\looseness-1

\input{tables/03-ablation_modalities}

\noindent \textbf{Choice of LLM architecture for Text Transformation:}
The choice of the pretrained LLM architecture plays a crucial role in improving the model performance of \tabglm. 
While larger LLMs like \cite{sun2023gpt, tabllm, tapex} ($\geq$7 billion parameters) can encode superior semantic features in complex text, it also adds a significant computational overhead. Additionally, their benefits may be negligible when dealing with simpler semantic content.
To address this trade off, we conducted an ablation experiment by varying the architecture of the text encoder ($E_{text}$) across three popular LLM models - TAPAS~\cite{tapas}, TAPEX~\cite{tapex} and TabLLM~\cite{tabllm}.
For all three settings we adopt the complete multi-modal training strategy, modifying only the text encoder $E_{text}$. 
The results from this experiment, shown in Table \ref{tab:choice_of_llm}, highlight that TAPAS~\footnote{We adopt the TAPAS-base model from \url{https://huggingface.co/google/tapas-base}}, a smaller parameter count, BERT~\cite{devlin2018bert} based text encoder, outperforms other larger models like TAPEX~\cite{tapex}. We thus adopt this architecture for the text transformation pipeline in \tabglm.\looseness-1

%% file: tables/01-benchmark_Sota_methods.tex
\begin{table*}[ht]
    \centering
    \scriptsize
    \begin{tabular}{l|cccccc|ccc}
    \toprule
    \multirow{2}{*}{\textbf{Dataset}} & \multicolumn{9}{c}{\textbf{Performance (AUC-ROC)}} \\
    \cline{2-10}
             & \textbf{\tabglm} & \textbf{CatBoost} & \textbf{GB} & \textbf{LR} & \textbf{RF} & \textbf{XGBoost} & \textbf{Tab} & \textbf{FT-} & \textbf{NODE} \\
             & (ours) &   &  &   &  &  & \textbf{Transformer} & \textbf{Transformer} &  \\
    \midrule \midrule
    bank         & 92.07 & \textbf{93.51}  & 92.36 & 86.76  & 92.46 & 92.84 & 90.05 &  92.07 & \underline{92.67} \\
    blood        & \textbf{78.48} & 74.94 & 72.24 & \underline{76.76} & 70.77 & 69.51 & 74.26 & 74.98 & 76.21 \\
    calhousing   & \textbf{95.47} & \underline{93.55} & 92.47 & 90.84 & 93.45  & 81.99 & 83.13 & 93.62 & 93.84 \\
    car          & 99.40 & \textbf{99.97} & \underline{99.83} & 78.46 & 99.41 & 99.92 & 98.57 & 98.51 & 99.64 \\
    coil2000     & \underline{74.17} & 73.97 & \textbf{74.66} & 73.22 & 69.43 & 71.19 & 71.64 & 65.59 & 73.09 \\
    creditg      & 79.32 & \textbf{80.54} & 78.36 & 75.21 & 79.76 & 76.81 & 79.40 & 56.60 & \underline{79.83} \\
    diabetes     & \textbf{83.70} & \underline{82.55} & 82.34 & 82.89 & 81.65 & 79.17 & 82.72 & 82.34 & 82.18 \\
    heart        & \textbf{93.29} & \underline{92.61} & 92.00 & 90.74 & 91.92 & 91.16 & 92.16 & 91.81 & \underline{92.61} \\
    kr-vs-kp     & \underline{99.43} & \textbf{99.95} & 99.77 & 99.15 & 99.86 & 99.95  & 99.30 & 86.79 & 99.41 \\
    mfeat-fourier & 99.94 & \underline{99.97} & 99.62 & \textbf{100.00} & 99.99 & 99.70 & 99.99 & 99.92 & \textbf{100.00} \\
    pc3          & \textbf{82.82} & \underline{82.48} & 80.80 & 79.44 & 80.89 & 77.76 & 79.02 & 76.57 & 81.00 \\
    income       & \textbf{92.59} & 92.44 & 91.75 & 79.03 & 89.19 & \underline{92.35} & 89.63 & 70.57 & 90.30 \\
    texture      & \textbf{100.0} & \underline{99.98} & 99.93 & 99.87 & 99.94 & 99.96 & 99.98 & 99.94 & 99.94 \\
    balance-scale       & \textbf{99.10} & 92.35 & \underline{98.37} & 93.11 & 84.89 & 98.99 & 91.60 & 91.03 & 94.41 \\
    mfeat-karhunen      & \textbf{99.88} & \underline{99.86} & 99.79 & 99.52 & 99.71 & 98.69 & 99.56 & 98.85 & \textbf{99.88} \\
    mfeat-morphological & \textbf{96.99} & 96.20 & 96.01 & 95.74 & 95.53 & 96.12 & 95.75 & 96.33 & \underline{96.34} \\
    mfeat-zernike       & \textbf{98.09} & \underline{97.59} & 97.16 & 97.74 & 96.72 & 97.35 & 98.02 & 97.76 & 97.49 \\
    cmc                 & \textbf{74.45} & 72.56 & \underline{72.89} & 70.41 & 70.52 & 73.00 & 69.96 & 71.56 & \underline{73.88} \\
    tic-tac-toe         & \underline{99.85} & 99.92 & 99.81 & 72.00 & 96.12 & \textbf{99.98} & 70.90 & 72.76 & 98.82 \\
    vehicle             & \textbf{94.50} & \underline{93.02} & 92.33 & 88.79 & 93.23 & 92.84 & 93.19 & 90.50 & 91.61 \\
    eucalyptus          & \textbf{91.95} & \underline{88.59} & 89.31 & 87.45 & 90.11 & 90.04 & 88.27 & 89.98 & 89.70 \\
    analcatdata\_author  & \textbf{58.96} & \underline{55.89} & 54.61 & 53.56 & 53.20 & 57.43 & 53.63 & 53.94 & 55.50 \\
    MiceProtein         & \underline{99.98} & \textbf{99.99} & 99.97 & 99.51 & 99.85 & \underline{99.98} & 99.91 & 99.41 & 99.97 \\
    steel-plates-fault  & 94.52 & \underline{96.51} & \textbf{96.26} & 91.35 & 91.71 & 96.56 & 91.91 & 91.92 & 94.45 \\
    dress-sales         & \textbf{57.89} & \underline{56.96} & 55.93 & 55.94 & 53.72 & 57.23 & 53.38 & 54.41 & 52.62 \\ \midrule
    \textbf{Average} & \textbf{89.47} & 88.64 & 88.34 & 84.69 & \underline{86.96} & 87.62 & 85.84 & 83.91 & 88.22 \\
    \bottomrule
    \end{tabular}
    \caption{\textbf{Comparison of performance (AUC-ROC) of existing approaches in tabular Machine Learning against \tabglm}. Our proposed method \tabglm\ achieves significant performance gains across 25 classification datasets. The best performing model is in \textbf{bold} while the second best is \underline{underlined}.}
    \label{tab:sota_perf_contrast}
\end{table*}

%% file: tables/02-benchmark_dl_methods.tex
\begin{table}[t]
\centering
\scriptsize
    \begin{tabular}{l|cccc}
    \toprule
    \multirow{3}{*}{ \textbf{Dataset} } & \multicolumn{4}{c}{ \textbf{Tabular DL Methods} } \\ \cline{2-5}
                               & \textbf{\tabglm} & \textbf{IGNNet} & \textbf{TabLLM} & \textbf{TabPFN} \\
                               & (multi-modal) & (graph) & (text) &  \\
    \midrule \midrule
        bank         & 92.07 & 91.11 & 91.20 & 91.19 \\
        blood        & \textbf{78.48} & 74.09 & 74.03 & 77.01 \\
        calhousing   & \textbf{95.47} & 94.79 & 95.38 & 95.31 \\
        car          & 99.40 & 50.16 & \textbf{99.99} & 99.53 \\
        creditg      & 79.32 & 71.99 & 70.82 & \textbf{80.79} \\
        diabetes     & \textbf{83.70} & 77.79 & 80.40 & 73.67 \\
        heart        & 93.29 & 92.06 & \textbf{94.21} & 82.60 \\
        jungle       & 88.98 & 88.98 & 93.00 & 87.36 \\
        income       & \textbf{92.59} & 90.76 & 92.19 & 90.14 \\ \midrule
        \textbf{Average} & \textbf{89.26} & 81.30 & 87.91 & 86.40 \\ \hline
    \end{tabular}
\caption{\textbf{Comparison of performance (AUC-ROC) of \tabglm\ against benchmark datasets in TabLLM} \cite{tabllm}. Results from all methods are averaged over five seeds.}
\label{tab:dl_model_benchmark}
\end{table}

%% file: tables/04-ablation_choice_of_llm.tex
\begin{table}[t]
\centering
\scriptsize
\begin{tabular}{l|ccc}
\toprule
\multirow{3}{*}{ \textbf{Dataset} } & \multicolumn{3}{c}{ \textbf{Methods} } \\ \cline{2 - 4}
                           & \multirow{2}{*}{ \textbf{TabLLM} } &  \textbf{\tabglm} & \textbf{\tabglm} \\
                                    &  & (w TAPEX encoder) & (w TAPAS encoder) \\
\midrule 
{Param. Count}& 2.9B& 336M & 129M\\
\midrule
blood              &  71.78   &   77.57   &  \textbf{78.48} \\
calhousing         &  95.00   &   95.29   &  \textbf{95.47} \\
creditg            &  78.56   &   78.72   &  \textbf{79.32} \\
\bottomrule
\end{tabular}
\caption{\textbf{Ablation on the Choice of LLM} architecture for the text transformation module of \tabglm.}
\label{tab:choice_of_llm}
\end{table}

%% file: tables/03-ablation_modalities.tex
\begin{table}[t]
      \centering
      \scriptsize
        \begin{tabular}{l|cc|c}
            \toprule
            \multirow{2}{*}{\textbf{Dataset}} & \textbf{Graph Trans.} & \textbf{Text Trans.} & \multirow{2}{*}{\textbf{AUCROC}} \\              
                                    &  ($E_{graph}$) & ($E_{text}$) &         \\
            \midrule \midrule
            \multirow{3}{*}{pc3}    & \checkmark  &           &  77.04 \\
                                    &            & \checkmark &  78.24 \\
                                    & \checkmark & \checkmark &  \textbf{82.82} \\
            \hline
            \multirow{3}{*}{bank }  & \checkmark &            &  91.11 \\
                                    &            & \checkmark &  90.52\\
                                    & \checkmark & \checkmark &  \textbf{92.07} \\
            \hline
            \multirow{3}{*}{Creditg}& \checkmark &            &  71.99 \\
                                    &            & \checkmark &  77.36\\
                                    & \checkmark & \checkmark &  \textbf{79.32} \\
            \bottomrule
      \end{tabular}
      \caption{\textbf{Ablations on the graph and text components of the proposed \tabglm\ approach}. Results are averaged over five seeds. }
      \label{tab:tabglm_components}
\end{table}

%% file: sections/06-conclusion.tex
In conclusion, \tabglm\ marks a pivotal advancement in deep learning for tabular data by adeptly handling the inherent heterogeneity of these datasets. 
By transforming each row into both a fully connected graph and serialized text, and leveraging a graph neural network alongside a pretrained text encoder, \tabglm\ captures rich structural and semantic information. 
Its joint multi-modal, semi-supervised learning objective enhances generalization and feature representation. 
The model's flexible graph-text pipeline efficiently processes diverse feature types, resulting in a streamlined architecture with significantly fewer parameters than state-of-the-art approaches. 
Evaluations across 25 benchmark datasets reveal substantial performance gains in AUC-ROC scores, with \tabglm\ surpassing both existing deep learning and traditional machine learning methods. 
These findings underscore the power of multi-modal architectures for tabular data, opening new horizons for innovative applications across various domains.\looseness-1

%% file: sections/07-appendix.tex
\section{Notations}
\label{app:notations}
In this section we pen down explanations for the symbols used in various equations and mathematical formualations in \tabglm.

\begin{table*}[t]
      \caption{Collection of notations used in \tabglm.}
      \centering
      \begin{tabular}{ c | c }
            \toprule
           \textbf{Symbol}  & \textbf{Description} \\
            \midrule
            $\mathcal{T}$ & Tabular dataset where $\mathcal{T} \in \mathbb{R}^{m \times n}$ with $|\mathcal{T}| = n$. \\
            $\mathcal{T}_{train}$ & Training dataset where $\mathcal{T}_{train} \subseteq \mathcal{T}$. \\
            $\mathcal{T}_{test}$ & Testing dataset where $\mathcal{T}_{test} \subseteq \mathcal{T}$. \\
            $n$ & Total number of rows in the dataset $\mathcal{T}$. \\
            $m$ & Total number of features (columns) in the dataset $\mathcal{T}$. \\ 
            $h(x, \theta)$ & Multi-Modal Neural Network used as feature extractor which accepts as $x$ (one row) as input. \\
            $Clf(.,.)$ & Multi-Layer Perceptron as classifier or Regressor. \\
            $\theta$ & Total Parameters of the feature extractor $h$ including both text and graph encoders. \\
            $\theta_{graph}$ & Parameters of the graph feature extractor (GNN in \cite{ignnet}). \\
            $\theta_{text}$ & Parameters of the text encoder (TAPAS~\cite{tapas}). \\
            $E_{graph}$ & Graph Encoder. Represents a Message-Passing based graph neural network~\cite{xu2019gnn}. \\
            $E_{text}$ & Text Encoder. Represented through a Large Language Model (LLM). \\
            $q_i$ & Intermediate representation of each row $\mathcal{T}_i$ represented as grpah $G_i$. \\
            $g_i^{graph}$ & Graph embedding corresponding to each row of the input table. \\
            $g_i^{text}$ & Text embedding corresponding to each row of the input table. \\
            $T_i^{ser}$ & Serialized text representation of each row of the input table. \\
            \texttt{proj} & MLP projection layer to bridge the gap between text and graph domains. \\ \bottomrule
      \end{tabular}\\
      \label{tab:notations}
      
\end{table*}

\section{Additional Setup Details}
\label{app:setup}
In this section we detail out additional setup details with details of hyper-parameters and outline of the software framework in \tabglm. 
At first, we detail out the Data-Preprocessing pipeline in \tabglm\ explained in Section 3.2 of the main paper.
The multi-modal pipeline in \tabglm\ transforms the input table $\mathcal{T}$ into graph and text representation. 
To minimize the sample level consistency in \tabglm, \textbf{each row in $\mathcal{T}$ is simultaneously converted to a serialized text $T_i^{ser}$ and a graph $G_i$}. 
The text serialization process is largely dependent on the choice of the tokenizer. In \tabglm\ we experiment on two different choices - TAPEX~\citep{tapex} and TAPAS~\citep{tapas} and the best performing architecture is chosen for downstream evaluation tasks (in our case this is TAPAS). Thus, we follow the serialization for TAPAS as shown in Figure 3 of the main paper.
The challenge lies in the graph transformation where existing methods like IGNNet~\citep{ignnet}, Table2Graph~\citep{huang22table2graph} etc. have assumed the input data to be numeric during the graph generation phase.
To overcome this challenge we convert the textual (categorical) columns into one-hot encodings, rendering them as numeric which are then joined to the existing numeric columns to form the final table for graph generation. Although, one-hot encoding adds additional columns to the input table $\mathcal{T}$, contrasting against other encoding techniques popular in Tabular DL literature like LabelEncoder or OrdinalEncoder from Scikit-learn, One-hot encoding produced the best downstream performance on several datasets. Thus we adopted On-hot encoding in our design of \tabglm. We ablate on the choice of encoders (LabelEncoder vs. One-Hot Encoder) in Table \ref{tab:choice_of_encoder} and conclude that One-Hot Encoding produces the best overall performance in \tabglm\ across 4 diverse datasets.

\begin{table}[h]
\centering
\caption{\textbf{Contrasting Encoding Strategies} for transforming heterogeneous tables to numeric ones for graph generation. Results from all methods are averaged over five seeds.}
    \begin{tabular}{l|cc}
    \toprule
    \multirow{3}{*}{ \textbf{Dataset} } & \multicolumn{2}{c}{ \textbf{Encoding Techniques} } \\ \cline{2 - 3}
                               & \textbf{\tabglm} & \textbf{\tabglm} \\
                               & (LabelEncoder) & (One-Hot Encoder)\\
    \midrule \midrule
        blood        & 78.29 & \textbf{78.49}  \\
        calhousing   & 95.34 & \textbf{95.47} \\
        coil2000     & 73.68          & \textbf{74.17} \\
        diabetes     & 83.14 & \textbf{83.70} \\ \midrule
        \textbf{Average} & 82.61 & \textbf{82.95} \\ \hline
    \end{tabular}
\label{tab:choice_of_encoder}
\end{table}

Since, our experiments consisted of several benchmark datasets spanning over multiple types of downstream tasks in classification and regression, it is challenging to tune the hyper-parameters for each and every dataset. We thus adopt a fixed set of hyper-parameters across every experiment in \tabglm. We adopt a fixed batch size of 256 for all benchmark datasets, with an initial learning rate of 1e-4 and weight the consistency loss in \textsc{MuCosa} at 0.2 which provides a higher weightage to the supervised loss allowing the model to rapidly adapt to downstream tasks. 
Additionally, we also introduce early stopping with a maximum number of training epochs set at 240. This paves the way for a fairer comparison across methods and datasets in \tabglm. Finally, since the performance of \tabglm\ varies with machine seed values we report the performance of each method across datasets on 5 fixed seeds - 5, 108, 180, 234 and 250. A particular seed value is set at the start of each experiment.
With the aim for introducing a framework for experimenting on tabular datasets we encapsulate the codebase for the aforementioned experiments with additional plug and play features for integrating new datasets and benchmarks, on \url{https://anonymous.4open.science/r/TabGLM/}.

\subsection{Dataset Details}
\label{sec:dataset_details}
As already summarized in Section 4.1 in the main paper, we conduct our benchmark experiments on 25 datasets spanning binary classification, multi-class classification and regression tasks. 
Given below are brief descriptions of each dataset describing the heterogeneity of the feature columns. A summary of the statistics corresponding to each dataset is also demonstrated in Table \ref{tab:dataset_details}. 

\begin{enumerate}
    \item \textbf{bank}\footnote{\url{https://archive.ics.uci.edu/ml/datasets/bank+marketing}}: The Bank Marketing dataset contains 7 numerical and 9 categorical columns, with 45,211 rows. It captures data related to a direct marketing campaign of a Portuguese bank, used to predict whether a client will subscribe to a term deposit.

    \item \textbf{blood}\footnote{\url{https://archive.ics.uci.edu/ml/datasets/Blood+Transfusion+Service+Center}}: This dataset includes 4 numerical columns and 748 rows, capturing features of blood donors used to predict whether a donor gave blood in a recent period.

    \item \textbf{calhousing}\footnote{\url{https://scikit-learn.org/stable/modules/generated/sklearn.datasets.fetch_california_housing.html}}: The California Housing dataset, consisting of 8 numerical columns and 20,640 rows, provides median house prices for California districts, commonly used for regression tasks. However, following the dataset specifications in TabLLM~\citep{tabllm} benchmark we convert this dataset to a binary classification setting with the positive label being assigned to properties with price $\geq$ median price.

    \item \textbf{car}\footnote{\url{https://archive.ics.uci.edu/ml/datasets/Car+Evaluation}}: The Car Evaluation dataset contains 6 categorical columns and 1,728 rows, assessing the quality of a car based on several criteria, used for multi-class classification.

    \item \textbf{coil2000}\footnote{\url{https://www.openml.org/d/384}}: This dataset comprises 86 numerical columns and 9,822 rows, containing information about clients of an insurance company, used to predict whether a client will be interested in a product.

    \item \textbf{creditg}\footnote{\url{https://archive.ics.uci.edu/ml/datasets/statlog+(german+credit+data)}}: The German Credit dataset includes 7 numerical and 13 categorical columns, with 1,000 rows. It is used to predict credit risk, making it a staple for binary classification tasks.

    \item \textbf{diabetes}\footnote{\url{https://www.kaggle.com/datasets/uciml/pima-indians-diabetes-database}}: The Pima Indians Diabetes dataset contains 8 numerical columns and 768 rows, capturing health parameters used to predict the onset of diabetes, widely used for binary classification.

    \item \textbf{heart}\footnote{\url{https://archive.ics.uci.edu/ml/datasets/heart+Disease}}: This dataset includes 6 numerical and 5 categorical columns, with 918 rows. It is used to predict heart disease, a common binary classification task.


    \item \textbf{kr-vs-kp}\footnote{\url{https://archive.ics.uci.edu/ml/datasets/kr-vs-kp}}: The King-Rook vs. King-Pawn dataset contains 37 numerical columns and 3,196 rows, with features extracted from chess endgames, used to classify game outcomes.

    \item \textbf{mfeat-fourier}\footnote{\url{https://archive.ics.uci.edu/ml/datasets/Multiple+Features}}: This dataset provides 77 numerical columns and 2,000 rows, consisting of Fourier coefficients of digitized images of handwritten digits, used for multi-class classification.

    \item \textbf{pc3}\footnote{\url{http://promise.site.uottawa.ca/SERepository/datasets-page.html}}: The PC3 dataset contains 38 numerical columns and 1,563 rows, from software defect prediction, used to classify software modules as defective or not.

    \item \textbf{income}\footnote{\url{https://archive.ics.uci.edu/ml/datasets/adult}}: The Adult dataset includes 4 numerical and 8 categorical columns, with 32,561 rows. It is often used for income prediction based on census data, making it a common binary classification task.

    \item \textbf{texture}\footnote{\url{https://archive.ics.uci.edu/ml/datasets/Multiple+Features}}: This dataset consists of 40 numerical and 1 categorical column, with 5,500 rows. It captures texture images, commonly used in image classification tasks.

    \item \textbf{balance-scale}\footnote{\url{https://archive.ics.uci.edu/ml/datasets/Balance+Scale}}: The Balance Scale dataset simulates the balance state of a scale based on the weight and distance from the center, with 5 numerical and 1 categorical column, totaling 625 rows, used for multi-class classification.

    \item \textbf{mfeat-karhunen}\footnote{\url{https://archive.ics.uci.edu/ml/datasets/Multiple+Features}}: This dataset contains 65 numerical and 1 categorical column, with 2,000 rows, capturing Karhunen-Loeve coefficients from digitized images of handwritten digits, used for multi-class classification.

    \item \textbf{mfeat-morphological}\footnote{\url{https://archive.ics.uci.edu/ml/datasets/Multiple+Features}}: This dataset is derived from morphological features of handwritten digits, containing 7 numerical and 1 categorical column, with 2,000 rows, used for multi-class classification tasks.

    \item \textbf{mfeat-zernike}\footnote{\url{https://archive.ics.uci.edu/ml/datasets/Multiple+Features}}: This dataset provides 48 numerical and 1 categorical column, with 2,000 rows, consisting of Zernike moments of digitized images of handwritten digits, used for multi-class classification tasks.

    \item \textbf{cmc}\footnote{\url{https://archive.ics.uci.edu/ml/datasets/Contraceptive+Method+Choice}}: The Contraceptive Method Choice dataset includes 10 numerical and 8 categorical columns, with 1,473 rows. It is used to predict contraceptive method choices among women, making it a common multi-class classification task.

    \item \textbf{tic-tac-toe}\footnote{\url{https://archive.ics.uci.edu/ml/datasets/Tic-Tac-Toe+Endgame}}: This dataset represents the states of a Tic-Tac-Toe game with 10 numerical and 10 categorical columns, totaling 690 rows, used to predict win/loss outcomes in a binary classification task.

    \item \textbf{vehicle}\footnote{\url{https://archive.ics.uci.edu/ml/datasets/Statlog+(Vehicle+Silhouettes)}}: This dataset contains 19 numerical and 1 categorical column, with 846 rows, used to classify different types of vehicles based on their silhouette, a common multi-class classification task.

    \item \textbf{eucalyptus}\footnote{\url{https://openml.org/d/188}}: The Eucalyptus dataset includes 20 numerical and 6 categorical columns, with 736 rows. It is used for the classification of eucalyptus species, making it a multi-class classification task.

    \item \textbf{analcatdata\_author}\footnote{\url{https://www.openml.org/d/183}}: The dataset includes 5 numerical and 5 categorical columns, with 797 rows, used for predicting the gender of an author based on bibliometric data, typically for binary classification.

    \item \textbf{MiceProtein}\footnote{\url{https://archive.ics.uci.edu/ml/datasets/Mice+Protein+Expression}}: This dataset consists of 82 numerical and 5 categorical columns, with 1,080 rows. It captures protein expression levels across multiple protein types, used for multi-class classification tasks.

    \item \textbf{steel-plates-fault}\footnote{\url{https://archive.ics.uci.edu/ml/datasets/Steel+Plates+Faults}}: This dataset contains 28 numerical and 1 categorical column, with 1,941 rows, capturing features related to faults in steel plates, used for multi-class classification of fault types.

    \item \textbf{dress-sales}\footnote{\url{https://www.openml.org/search?type=data&status=active&id=23381}}: This dataset includes 13 numerical and 12 categorical columns, with 500 rows, containing sales data for different dress designs, used for binary classification tasks as used in OpenML.

\end{enumerate}

\begin{table}[h]
    \centering
    \caption{Adaptation of \tabglm\ towards categorical datasets by employing only the text pipeline during model training. The results are averaged over 5 seeds on the \textbf{car} dataset.}
    \begin{tabular}{c|c}
         \toprule
         \textbf{Design Strategy in \tabglm} & AUC-ROC \\ \midrule
         \tabglm\ Full (multi-modal)    & 98.37  \\
         \tabglm\ text-only (uni-modal) & \textbf{99.40}  \\
         \bottomrule
    \end{tabular}    
    \label{tab:text_only}
\end{table}

\begin{table*}[ht]
    \centering
    \caption{\textbf{Detailed Statistics of Datasets in \tabglm\ benchmark.}}
    \label{tab:dataset_details}
    \resizebox{0.8\textwidth}{!}{
        \begin{tabular}{l|c|ccc}
        \toprule
        \textbf{Dataset} & \textbf{Type} & \textbf{Numerical Columns} & \textbf{Categorical Columns} & \textbf{No. of Rows} \\
        \midrule \midrule
        bank             & Binary        & 7   & 9  & 45,211 \\
        blood            & Binary        & 4   & 0  & 748 \\
        calhousing       & Binary\footnote{As in TabLLM~\citep{tabllm}}    & 8   & 0  & 20,640 \\
        car              & Multi-class   & 0   & 6  & 1,728 \\
        coil2000         & Binary        & 86  & 0  & 9,822 \\
        creditg          & Binary        & 7   & 13 & 1,000 \\
        diabetes         & Binary        & 8   & 0  & 768 \\
        heart            & Binary        & 6   & 5  & 918 \\
        kr-vs-kp         & Binary        & 37  & 0  & 3,196 \\
        mfeat-fourier    & Multi-class   & 77  & 0  & 2,000 \\
        pc3              & Binary        & 38  & 0  & 1,563 \\
        income           & Binary        & 4   & 8  & 32,561 \\
        texture          & Multi-class   & 40  & 1  & 5,500 \\
        balance-scale    & Multi-class   & 5   & 1  & 625 \\
        mfeat-karhunen   & Multi-class   & 65  & 1  & 2,000 \\
        mfeat-morphological & Multi-class & 7   & 1  & 2,000 \\
        mfeat-zernike    & Multi-class   & 48  & 1  & 2,000 \\
        cmc              & Multi-class   & 10  & 8  & 1,473 \\
        tic-tac-toe      & Binary        & 10  & 10 & 690 \\
        vehicle          & Multi-class   & 19  & 1  & 846 \\
        eucalyptus       & Multi-class   & 20  & 6  & 736 \\
        analcatdata\_author & Binary      & 5   & 5  & 797 \\
        MiceProtein      & Multi-class   & 82  & 5  & 1,080 \\
        steel-plates-fault & Multi-class & 28  & 1  & 1,941 \\
        dress-sales      & Binary        & 1  & 12 & 500 \\
        \bottomrule
        \end{tabular}
    }
\end{table*}

\subsection{Additional Explanations to Results in Table 1}
Following the discussion in Section 4.4 we show in Table 1, the performance of our approach \tabglm\ on 25 diverse benchmarks spanning varied dataset sizes and degree of heterogeneity (number of numerical vs. categorical columns in the table). 
For each dataset we run each method for 5 distinct seed values and report the average AUC-ROC scores in Table 1 with standard deviations provided in Table \ref{tab:sota_perf_contrast_with_std}. 
On an average we show that \tabglm\ outperforms both traditional ML models and newer Tabular DL models on most downstream tasks. In this section, we provide a more granular analysis of our results.
At first, we observe that \tabglm\ consistently outperforms existing approaches in datasets with at least 1 heterogeneous feature column demonstrating its capability to learn discriminative feature representations from heterogeneous datasets. 
Secondly, we observe that \tabglm\ consistently outperforms recent DL based approaches like NODE~\citep{popov2019neural}, TabTransformer~\citep{huang2020tabtransformer} and unimodal approaches like IGNNet and TabLLM~\citep{tabllm} by significant margins. This is particularly due to the fact that DL based approaches model either spatial, semantic or structural features completely missing the importance of modelling the benefits of auxiliary modalities. 
Thirdly, we observe that for numeric tables like \textbf{blood}, \textbf{calhousing} graph based models like IGNNet provide significant boost in performance either comparable or greater than that of \tabglm\ (refer Table 3). On the contrary for categorical tables (containing only categorical columns), IGNNet significantly underperforms over \tabglm\ highlighting the importance of multi-modal learning to address data heterogeneity.

\begin{table*}[ht]
\centering
\scriptsize
\caption{\textbf{Comparison of performance (AUCROC) including Standard Deviations (Std) of existing approaches in tabular Machine Learning against \tabglm}. Our proposed method \tabglm\ achieves significant performance gains across 25 classification datasets. The best performing model is highlighted in bold, while the second best is italicized.}
\label{tab:sota_perf_contrast_with_std}
\resizebox{\textwidth}{!}{
\begin{tabular}{l|rl|rl|rl|rl|rl|rl|rl|rl|rl}
\toprule
& \multicolumn{2}{c}{\textbf{TabGLM (ours)}} & \multicolumn{2}{c}{\textbf{CatBoost}} & \multicolumn{2}{c}{\textbf{GB}} & \multicolumn{2}{c}{\textbf{LR}} & \multicolumn{2}{c}{\textbf{RF}} & \multicolumn{2}{c}{\textbf{XGBoost}} & \multicolumn{2}{c}{\textbf{TabTransformer}} & \multicolumn{2}{c}{\textbf{FT-Transformer}} & \multicolumn{2}{c}{\textbf{NODE}}\\
\cmidrule(lr){2-3} \cmidrule(lr){4-5} \cmidrule(lr){6-7} \cmidrule(lr){8-9}
\cmidrule(lr){10-11} \cmidrule(lr){12-13} \cmidrule(lr){14-15}
\cmidrule(lr){16-17} \cmidrule(lr){18-19}
Dataset & AUC-ROC & Std & AUC-ROC & Std & AUC-ROC & Std & AUC-ROC & Std
        & AUC-ROC & Std & AUC-ROC & Std & AUC-ROC & Std & AUC-ROC & Std
        & AUC-ROC & Std \\
\midrule
\midrule
bank                 & 92.07 & $\pm$0.10 & 93.51 & $\pm$0.03 & 92.36 & $\pm$0.01 & 86.76 & $\pm$0.11 & 92.46 & $\pm$0.01 & 92.84 & $\pm$0.03 & 90.05 & $\pm$0.06 & 92.07 & $\pm$0.06 & 92.67 & $\pm$0.01 \\
blood                & 78.48 & $\pm$0.04 & 74.94 & $\pm$0.01 & 72.24 & $\pm$0.01 & 76.76 & $\pm$0.02 & 70.77 & $\pm$0.01 & 69.51 & $\pm$0.01 & 74.26 & $\pm$0.04 & 74.98 & $\pm$0.10 & 76.21 & $\pm$0.06 \\
calhousing           & 95.47 & $\pm$0.03 & 93.55 & $\pm$0.03 & 92.47 & $\pm$0.03 & 90.84 & $\pm$0.02 & 93.45 & $\pm$0.02 & 81.99 & $\pm$0.01 & 83.13 & $\pm$0.13 & 93.62 & $\pm$0.07 & 93.84 & $\pm$0.05 \\
car                  & 99.40 & $\pm$0.04 & 99.97 & $\pm$0.02 & 99.83 & $\pm$0.02 & 78.46 & $\pm$1.03 & 99.41 & $\pm$0.15 & 99.92 & $\pm$0.08 & 98.57 & $\pm$0.11 & 98.51 & $\pm$0.20 & 99.64 & $\pm$0.27 \\
coil2000             & 74.17 & $\pm$0.40 & 73.97 & $\pm$1.20 & 74.66 & $\pm$0.60 & 73.22 & $\pm$1.10 & 69.43 & $\pm$1.21 & 71.19 & $\pm$1.00 & 71.64 & $\pm$1.00 & 65.59 & $\pm$0.34 & 73.09 & $\pm$0.12 \\
creditg              & 79.32 & $\pm$0.10 & 80.54 & $\pm$0.06 & 78.36 & $\pm$0.03 & 75.21 & $\pm$0.10 & 79.76 & $\pm$0.06 & 76.81 & $\pm$0.02 & 79.40 & $\pm$0.10 & 56.60 & $\pm$0.16 & 79.83 & $\pm$0.08 \\
diabetes             & 83.70 & $\pm$0.03 & 82.55 & $\pm$0.03 & 82.34 & $\pm$0.02 & 82.89 & $\pm$0.02 & 81.65 & $\pm$0.03 & 79.17 & $\pm$0.03 & 82.72 & $\pm$0.07 & 82.34 & $\pm$0.08 & 82.18 & $\pm$0.02 \\
heart                & 93.29 & $\pm$0.01 & 92.61 & $\pm$0.01 & 92.00 & $\pm$0.01 & 90.74 & $\pm$0.02 & 91.92 & $\pm$0.01 & 91.16 & $\pm$0.01 & 92.16 & $\pm$0.02 & 91.81 & $\pm$0.02 & 92.61 & $\pm$0.01 \\
kr-vs-kp             & 99.43 & $\pm$0.15 & 99.95 & $\pm$0.50 & 99.77 & $\pm$0.20 & 99.15 & $\pm$1.46 & 99.86 & $\pm$0.35 & 99.95 & $\pm$0.30 & 99.30 & $\pm$0.71 & 86.79 & $\pm$0.88 & 99.41 & $\pm$0.97 \\
mfeat-fourier        & 99.94 & $\pm$1.10 & 99.97 & $\pm$1.42 & 99.62 & $\pm$1.70 & 100.00 & $\pm$1.50 & 99.99 & $\pm$1.05 & 99.70 & $\pm$1.05 & 99.99 & $\pm$1.75 & 99.92 & $\pm$1.74 & 100.00 & $\pm$1.06 \\
pc3                  & 82.82 & $\pm$1.24 & 82.48 & $\pm$0.65 & 80.80 & $\pm$2.71 & 79.44 & $\pm$1.79 & 80.89 & $\pm$3.37 & 77.76 & $\pm$3.32 & 79.02 & $\pm$4.20 & 76.57 & $\pm$3.42 & 81.00 & $\pm$0.45 \\
income               & 92.59 & $\pm$0.13 & 92.44 & $\pm$0.04 & 91.75 & $\pm$0.03 & 79.03 & $\pm$0.09 & 89.19 & $\pm$0.02 & 92.35 & $\pm$0.10 & 89.63 & $\pm$0.08 & 70.57 & $\pm$0.10 & 90.30 & $\pm$0.05 \\
texture              & 100.00 & $\pm$0.44 & 99.98 & $\pm$0.27 & 99.93 & $\pm$0.30 & 99.87 & $\pm$0.53 & 99.94 & $\pm$0.26 & 99.96 & $\pm$0.27 & 99.98 & $\pm$1.16 & 99.94 & $\pm$1.07 & 99.94 & $\pm$0.70 \\
balance-scale        & 99.10 & $\pm$0.03 & 92.35 & $\pm$0.02 & 98.37 & $\pm$0.00 & 93.11 & $\pm$0.02 & 84.89 & $\pm$0.03 & 98.99 & $\pm$0.00 & 91.60 & $\pm$0.21 & 91.03 & $\pm$0.18 & 94.41 & $\pm$0.01 \\
mfeat-karhunen       & 99.88 & $\pm$0.56 & 99.86 & $\pm$0.44 & 99.79 & $\pm$0.47 & 99.52 & $\pm$1.43 & 99.71 & $\pm$0.40 & 98.69 & $\pm$0.32 & 99.56 & $\pm$1.22 & 98.85 & $\pm$1.19 & 99.88 & $\pm$0.38 \\
mfeat-morphological  & 96.99 & $\pm$0.37 & 96.20 & $\pm$0.42 & 96.01 & $\pm$0.30 & 95.74 & $\pm$1.25 & 95.53 & $\pm$0.31 & 96.12 & $\pm$0.30 & 95.75 & $\pm$1.01 & 96.33 & $\pm$1.01 & 96.34 & $\pm$0.33 \\
mfeat-zernike        & 98.09 & $\pm$0.42 & 97.59 & $\pm$0.35 & 97.16 & $\pm$0.30 & 97.74 & $\pm$0.51 & 96.72 & $\pm$0.30 & 97.35 & $\pm$0.34 & 98.02 & $\pm$0.65 & 97.76 & $\pm$0.68 & 97.49 & $\pm$0.53 \\
cmc                  & 74.45 & $\pm$1.22 & 72.56 & $\pm$1.26 & 72.89 & $\pm$1.62 & 70.41 & $\pm$1.44 & 70.52 & $\pm$1.10 & 73.00 & $\pm$1.73 & 69.96 & $\pm$1.66 & 71.56 & $\pm$1.64 & 73.88 & $\pm$1.10 \\
tic-tac-toe          & 99.85 & $\pm$0.01 & 99.92 & $\pm$0.01 & 99.81 & $\pm$0.00 & 72.00 & $\pm$0.06 & 96.12 & $\pm$0.01 & 99.98 & $\pm$0.01 & 70.90 & $\pm$0.07 & 72.76 & $\pm$0.09 & 98.82 & $\pm$0.03 \\
vehicle              & 94.50 & $\pm$0.04 & 93.02 & $\pm$0.08 & 92.33 & $\pm$1.07 & 88.79 & $\pm$0.11 & 93.23 & $\pm$0.07 & 92.84 & $\pm$0.06 & 93.19 & $\pm$0.18 & 90.50 & $\pm$0.21 & 91.61 & $\pm$0.13 \\
eucalyptus           & 91.95 & $\pm$0.55 & 88.59 & $\pm$0.73 & 89.31 & $\pm$0.54 & 87.45 & $\pm$1.30 & 90.11 & $\pm$0.82 & 90.04 & $\pm$1.12 & 88.27 & $\pm$1.43 & 89.98 & $\pm$1.71 & 89.70 & $\pm$1.04 \\
analcatdata\_author  & 58.96 & $\pm$0.30 & 55.89 & $\pm$0.41 & 54.61 & $\pm$0.37 & 53.56 & $\pm$0.55 & 53.20 & $\pm$0.40 & 57.43 & $\pm$0.35 & 53.63 & $\pm$0.50 & 53.94 & $\pm$0.62 & 55.50 & $\pm$0.36 \\
MiceProtein          & 99.98 & $\pm$0.01 & 99.99 & $\pm$0.00 & 99.97 & $\pm$0.02 & 99.51 & $\pm$0.02 & 99.85 & $\pm$0.00 & 99.98 & $\pm$0.01 & 99.91 & $\pm$0.04 & 99.41 & $\pm$0.10 & 99.97 & $\pm$0.01 \\
steel-plates-fault   & 94.52 & $\pm$0.05 & 96.51 & $\pm$0.06 & 96.26 & $\pm$0.04 & 91.35 & $\pm$0.04 & 91.71 & $\pm$0.06 & 96.56 & $\pm$0.03 & 91.91 & $\pm$0.12 & 91.92 & $\pm$0.10 & 94.45 & $\pm$0.06 \\
dress-sales          & 57.89 & $\pm$0.58 & 56.96 & $\pm$0.56 & 55.93 & $\pm$0.47 & 55.94 & $\pm$1.29 & 53.72 & $\pm$0.55 & 57.23 & $\pm$0.56 & 53.38 & $\pm$1.25 & 54.41 & $\pm$1.17 & 52.62 & $\pm$0.49 \\
\bottomrule
\end{tabular}}
\end{table*}

\subsection{Handling of Textual Only tables}
An important challenge in multi-modal tabular deep learning arises when the input dataset contains only categorical columns as features.
This renders the graph transformation pipeline ineffective in modelling structure of the underlying table. 
\tabglm\ addresses this by only adopting the text transformation pipeline (pretrained encoder in TAPAS alongside the classifier layers) in such situations significantly boosting performance.
This strategy is also adopted if there exists columns in the table with unique entries greater than 60\% of the number of rows in the table. A comparison between the full architecture of \tabglm\ and the text-only (adopting only the text transformation pipeline) architecture in \tabglm\ for the \textbf{car} dataset (only dataset in the benchmark with text only columns) has been given in Table \ref{tab:text_only}.

\section{Limitations}
\label{app:limitations}
Although \tabglm\ demonstrates significant developments over SoTA approaches demonstrating feature heterogeneity, but it also demonstrates some limitations. At first, the maximum number of columns that an input table can have is capped by the token limit of the underlying LLM requiring us to adopt larger networks with large memory and compute footprints. Secondly, the design of the graph transformation pipeline uses the column indexes and the statistical relationships between columns to learn the structure of the graph but does not consider the column name during node creation. These limitations of \tabglm\ would be handled in future research.

%% file: main_arxiv.bbl
\begin{thebibliography}{47}
\providecommand{\natexlab}[1]{#1}

\bibitem[{Alkhatib et~al.(2024)Alkhatib, Ennadir, Bostrom, and Vazirgiannis}]{ignnet}
Alkhatib, A.; Ennadir, S.; Bostrom, H.; and Vazirgiannis, M. 2024.
\newblock Interpretable Graph Neural Networks for Tabular Data.
\newblock In \emph{ICLR 2024 Workshop on Data-centric Machine Learning Research (DMLR): Harnessing Momentum for Science}.

\bibitem[{Arik and Pfister(2021)}]{arik2021tabnet}
Arik, S.~{\"O}.; and Pfister, T. 2021.
\newblock Tabnet: Attentive interpretable tabular learning.
\newblock In \emph{Proceedings of the AAAI conference on artificial intelligence}, volume~35, 6679--6687.

\bibitem[{Badaro, Saeed, and Papotti(2023)}]{badaro2023transformers}
Badaro, G.; Saeed, M.; and Papotti, P. 2023.
\newblock Transformers for Tabular Data Representation: A survey of models and applications.
\newblock \emph{Transactions of the Association for Computational Linguistics}, 11: 227--249.

\bibitem[{Bahri et~al.(2021)Bahri, Jiang, Tay, and Metzler}]{bahri2021scarf}
Bahri, D.; Jiang, H.; Tay, Y.; and Metzler, D. 2021.
\newblock Scarf: Self-supervised contrastive learning using random feature corruption.
\newblock \emph{arXiv preprint arXiv:2106.15147}.

\bibitem[{Bent{\'e}jac, Cs{\"o}rg{\H{o}}, and Mart{\'\i}nez-Mu{\~n}oz(2021)}]{bentejac2021comparative}
Bent{\'e}jac, C.; Cs{\"o}rg{\H{o}}, A.; and Mart{\'\i}nez-Mu{\~n}oz, G. 2021.
\newblock A comparative analysis of gradient boosting algorithms.
\newblock \emph{Artificial Intelligence Review}, 54: 1937--1967.

\bibitem[{Breiman(2001)}]{breiman2001random}
Breiman, L. 2001.
\newblock Random forests.
\newblock \emph{Machine learning}, 45: 5--32.

\bibitem[{Casalicchio et~al.(2017)Casalicchio, Bossek, Lang, Kirchhoff, Kerschke, Hofner, Seibold, Vanschoren, and Bischl}]{openml2017}
Casalicchio, G.; Bossek, J.; Lang, M.; Kirchhoff, D.; Kerschke, P.; Hofner, B.; Seibold, H.; Vanschoren, J.; and Bischl, B. 2017.
\newblock OpenML: An R package to connect to the machine learning platform OpenML.
\newblock \emph{Computational Statistics}, 32(3): 1--15.

\bibitem[{Chen et~al.(2023{\natexlab{a}})Chen, Xing, Wang, and Zhang}]{Chen_2023}
Chen, M.; Xing, L.; Wang, Y.; and Zhang, X. 2023{\natexlab{a}}.
\newblock Enhanced Multimodal Representation Learning with Cross-modal KD.
\newblock In \emph{2023 IEEE/CVF Conference on Computer Vision and Pattern Recognition (CVPR)}.

\bibitem[{Chen et~al.(2023{\natexlab{b}})Chen, Sarkar, Lausen, Srinivasan, Zha, Huang, and Karypis}]{chen2023hytrel}
Chen, P.; Sarkar, S.; Lausen, L.; Srinivasan, B.; Zha, S.; Huang, R.; and Karypis, G. 2023{\natexlab{b}}.
\newblock HyTrel: Hypergraph-enhanced Tabular Data Representation Learning.
\newblock In \emph{Thirty-seventh Conference on Neural Information Processing Systems}.

\bibitem[{Chen and Guestrin(2016)}]{chen2016xgboost}
Chen, T.; and Guestrin, C. 2016.
\newblock Xgboost: A scalable tree boosting system.
\newblock In \emph{Proceedings of the 22nd acm sigkdd international conference on knowledge discovery and data mining}, 785--794.

\bibitem[{Chen et~al.(2020)Chen, Kornblith, Norouzi, and Hinton}]{chen2020simple}
Chen, T.; Kornblith, S.; Norouzi, M.; and Hinton, G. 2020.
\newblock A simple framework for contrastive learning of visual representations.
\newblock In \emph{International conference on machine learning}, 1597--1607. PMLR.

\bibitem[{Devlin et~al.(2018)Devlin, Chang, Lee, and Toutanova}]{devlin2018bert}
Devlin, J.; Chang, M.-W.; Lee, K.; and Toutanova, K. 2018.
\newblock Bert: Pre-training of deep bidirectional transformers for language understanding.
\newblock \emph{arXiv preprint arXiv:1810.04805}.

\bibitem[{Galkin et~al.(2024)Galkin, Yuan, Mostafa, Tang, and Zhu}]{galkin2024towards}
Galkin, M.; Yuan, X.; Mostafa, H.; Tang, J.; and Zhu, Z. 2024.
\newblock Towards Foundation Models for Knowledge Graph Reasoning.
\newblock In \emph{The Twelfth International Conference on Learning Representations}.

\bibitem[{Geurts, Ernst, and Wehenkel(2006)}]{geurts2006extremely}
Geurts, P.; Ernst, D.; and Wehenkel, L. 2006.
\newblock Extremely randomized trees.
\newblock \emph{Machine learning}, 63: 3--42.

\bibitem[{Gorishniy et~al.(2021)Gorishniy, Rubachev, Khrulkov, and Babenko}]{gorishniy2021revisiting}
Gorishniy, Y.; Rubachev, I.; Khrulkov, V.; and Babenko, A. 2021.
\newblock Revisiting deep learning models for tabular data.
\newblock \emph{Advances in Neural Information Processing Systems}, 34: 18932--18943.

\bibitem[{Grinsztajn, Oyallon, and Varoquaux(2022)}]{grinsztajn2022tree}
Grinsztajn, L.; Oyallon, E.; and Varoquaux, G. 2022.
\newblock Why do tree-based models still outperform deep learning on typical tabular data?
\newblock \emph{Advances in Neural Information Processing Systems}, 35: 507--520.

\bibitem[{Guo et~al.(2021)Guo, Quan, Zhao, Yao, Li, and Tu}]{tabgnn}
Guo, X.; Quan, Y.; Zhao, H.; Yao, Q.; Li, Y.; and Tu, W.-W. 2021.
\newblock TabGNN: Multiplex Graph Neural Network for Tabular Data Prediction.
\newblock In \emph{DLP-KDD}.

\bibitem[{Hegde, Jose~Valanarasu, and Patel(2023)}]{hegde2023clip3d}
Hegde, D.; Jose~Valanarasu, J.~M.; and Patel, V.~M. 2023.
\newblock CLIP goes 3D: Leveraging Prompt Tuning for Language Grounded 3D Recognition.
\newblock In \emph{2023 IEEE/CVF International Conference on Computer Vision Workshops (ICCVW)}.

\bibitem[{Hegselmann et~al.(2023)Hegselmann, Buendia, Lang, Agrawal, Jiang, and Sontag}]{tabllm}
Hegselmann, S.; Buendia, A.; Lang, H.; Agrawal, M.; Jiang, X.; and Sontag, D. 2023.
\newblock Tabllm: Few-shot classification of tabular data with large language models.
\newblock In \emph{International Conference on Artificial Intelligence and Statistics}, 5549--5581. PMLR.

\bibitem[{Herzig et~al.(2020)Herzig, Nowak, M{\"u}ller, Piccinno, and Eisenschlos}]{tapas}
Herzig, J.; Nowak, P.~K.; M{\"u}ller, T.; Piccinno, F.; and Eisenschlos, J. 2020.
\newblock {T}a{P}as: Weakly Supervised Table Parsing via Pre-training.
\newblock In \emph{Proceedings of the 58th Annual Meeting of the Association for Computational Linguistics}.

\bibitem[{Hollmann et~al.(2023)Hollmann, M{\"u}ller, Eggensperger, and Hutter}]{hollmann2022tabpfn}
Hollmann, N.; M{\"u}ller, S.; Eggensperger, K.; and Hutter, F. 2023.
\newblock Tab{PFN}: A Transformer That Solves Small Tabular Classification Problems in a Second.
\newblock In \emph{The Eleventh International Conference on Learning Representations}.

\bibitem[{Hosmer~Jr, Lemeshow, and Sturdivant(2013)}]{hosmer2013applied}
Hosmer~Jr, D.~W.; Lemeshow, S.; and Sturdivant, R.~X. 2013.
\newblock \emph{Applied logistic regression}, volume 398.
\newblock John Wiley \& Sons.

\bibitem[{Huang et~al.(2020)Huang, Khetan, Cvitkovic, and Karnin}]{huang2020tabtransformer}
Huang, X.; Khetan, A.; Cvitkovic, M.; and Karnin, Z. 2020.
\newblock TabTransformer: Tabular Data Modeling Using Contextual Embeddings.
\newblock arXiv:2012.06678.

\bibitem[{Ke et~al.(2017)Ke, Meng, Finley, Wang, Chen, Ma, Ye, and Liu}]{ke2017lightgbm}
Ke, G.; Meng, Q.; Finley, T.; Wang, T.; Chen, W.; Ma, W.; Ye, Q.; and Liu, T.-Y. 2017.
\newblock Lightgbm: A highly efficient gradient boosting decision tree.
\newblock \emph{Advances in neural information processing systems}, 30.

\bibitem[{Kim, Grinsztajn, and Varoquaux(2024)}]{kim2024carte}
Kim, M.~J.; Grinsztajn, L.; and Varoquaux, G. 2024.
\newblock {CARTE}: Pretraining and Transfer for Tabular Learning.
\newblock In \emph{Proceedings of the 41st International Conference on Machine Learning}, volume 235 of \emph{PMLR}, 23843--23866.

\bibitem[{Liu et~al.(2022)Liu, Chen, Guo, Ziyadi, Lin, Chen, and Lou}]{tapex}
Liu, Q.; Chen, B.; Guo, J.; Ziyadi, M.; Lin, Z.; Chen, W.; and Lou, J.-G. 2022.
\newblock {TAPEX}: Table Pre-training via Learning a Neural {SQL} Executor.
\newblock In \emph{International Conference on Learning Representations}.

\bibitem[{Liu et~al.(2021)Liu, Zhang, Hou, Mian, Wang, Zhang, and Tang}]{liu2021self}
Liu, X.; Zhang, F.; Hou, Z.; Mian, L.; Wang, Z.; Zhang, J.; and Tang, J. 2021.
\newblock Self-supervised learning: Generative or contrastive.
\newblock \emph{IEEE transactions on knowledge and data engineering}, 35(1): 857--876.

\bibitem[{Majmundar et~al.(2022)Majmundar, Goyal, Netrapalli, and Jain}]{majmundar2022met}
Majmundar, K.; Goyal, S.; Netrapalli, P.; and Jain, P. 2022.
\newblock Met: Masked encoding for tabular data.
\newblock \emph{arXiv preprint arXiv:2206.08564}.

\bibitem[{Margeloiu et~al.(2023)Margeloiu, Simidjievski, Lio, and Jamnik}]{margeloiu2023gcondnet}
Margeloiu, A.; Simidjievski, N.; Lio, P.; and Jamnik, M. 2023.
\newblock {GC}ondNet: A Novel Method for Improving Neural Networks on Small High-Dimensional Tabular Data.
\newblock In \emph{NeurIPS 2023 Second Table Representation Learning Workshop}.

\bibitem[{Popov, Morozov, and Babenko(2019)}]{popov2019neural}
Popov, S.; Morozov, S.; and Babenko, A. 2019.
\newblock Neural oblivious decision ensembles for deep learning on tabular data.
\newblock \emph{arXiv preprint arXiv:1909.06312}.

\bibitem[{Prokhorenkova et~al.(2018)Prokhorenkova, Gusev, Vorobev, Dorogush, and Gulin}]{prokhorenkova2018catboost}
Prokhorenkova, L.; Gusev, G.; Vorobev, A.; Dorogush, A.~V.; and Gulin, A. 2018.
\newblock CatBoost: unbiased boosting with categorical features.
\newblock \emph{Advances in neural information processing systems}, 31.

\bibitem[{Radford et~al.(2021)Radford, Kim, Hallacy, Ramesh, Goh, Agarwal, Sastry, Askell, Mishkin, Clark et~al.}]{radford2021learning}
Radford, A.; Kim, J.~W.; Hallacy, C.; Ramesh, A.; Goh, G.; Agarwal, S.; Sastry, G.; Askell, A.; Mishkin, P.; Clark, J.; et~al. 2021.
\newblock Learning Transferable Visual Models From Natural Language Supervision.
\newblock \emph{arXiv preprint arXiv:2103.00020}.

\bibitem[{Ramesh et~al.(2021)Ramesh, Pavlov, Goh, Gray, Voss, Radford, Chen, and Sutskever}]{ramesh2021zero}
Ramesh, A.; Pavlov, M.; Goh, G.; Gray, S.; Voss, C.; Radford, A.; Chen, M.; and Sutskever, I. 2021.
\newblock Zero-shot text-to-image generation.
\newblock \emph{arXiv preprint arXiv:2102.12092}.

\bibitem[{Rubachev et~al.(2022)Rubachev, Alekberov, Gorishniy, and Babenko}]{rubachev2022revisiting}
Rubachev, I.; Alekberov, A.; Gorishniy, Y.; and Babenko, A. 2022.
\newblock Revisiting pretraining objectives for tabular deep learning.
\newblock \emph{arXiv preprint arXiv:2207.03208}.

\bibitem[{Sharma et~al.(2019)Sharma, Vans, Shigemizu, Boroevich, and Tsunoda}]{DeepInsight}
Sharma, A.; Vans, E.; Shigemizu, D.; Boroevich, K.~A.; and Tsunoda, T. 2019.
\newblock DeepInsight: A methodology to transform a non-image data to an image for convolution neural network architecture.
\newblock \emph{Scientific Reports}, 9: 11399.

\bibitem[{Shwartz-Ziv and Armon(2022)}]{shwartz2022tabular}
Shwartz-Ziv, R.; and Armon, A. 2022.
\newblock Tabular data: Deep learning is not all you need.
\newblock \emph{Information Fusion}, 81: 84--90.

\bibitem[{Somepalli et~al.(2021)Somepalli, Goldblum, Schwarzschild, Bruss, and Goldstein}]{somepalli2021saint}
Somepalli, G.; Goldblum, M.; Schwarzschild, A.; Bruss, C.~B.; and Goldstein, T. 2021.
\newblock Saint: Improved neural networks for tabular data via row attention and contrastive pre-training.
\newblock \emph{arXiv preprint arXiv:2106.01342}.

\bibitem[{Sun(2023)}]{sun2023gpt}
Sun, e.~a. 2023.
\newblock Graph Propagation Transformer for Graph Representation Learning.
\newblock \emph{Proceedings of the International Joint Conference on Artificial Intelligence (IJCAI)}.

\bibitem[{Ucar, Hajiramezanali, and Edwards(2021)}]{ucar2021subtab}
Ucar, T.; Hajiramezanali, E.; and Edwards, L. 2021.
\newblock Subtab: Subsetting features of tabular data for self-supervised representation learning.
\newblock \emph{Advances in Neural Information Processing Systems}, 34: 18853--18865.

\bibitem[{Wang et~al.(2019)Wang, Li, Gao, and Zhang}]{wang2019supertml}
Wang, P.; Li, K.; Gao, J.; and Zhang, C. 2019.
\newblock SuperTML: Two-Dimensional Word Embedding for the Precognition on Structured Tabular Data.
\newblock In \emph{Proceedings of the 25th ACM SIGKDD International Conference on Knowledge Discovery \& Data Mining}, 2327--2335.

\bibitem[{Wang and Sun(2022)}]{wang2022transtab}
Wang, Z.; and Sun, J. 2022.
\newblock Transtab: Learning transferable tabular transformers across tables.
\newblock \emph{Advances in Neural Information Processing Systems}, 35: 2902--2915.

\bibitem[{Wu et~al.(2021)Wu, Wu, Qi, Huang, and Xie}]{wu2021fastformer}
Wu, C.; Wu, F.; Qi, T.; Huang, Y.; and Xie, X. 2021.
\newblock Fastformer: Additive attention can be all you need.
\newblock \emph{arXiv preprint arXiv:2108.09084}.

\bibitem[{Xu et~al.(2019)Xu, Hu, Leskovec, and Jegelka}]{xu2019gnn}
Xu, K.; Hu, W.; Leskovec, J.; and Jegelka, S. 2019.
\newblock How Powerful are Graph Neural Networks?
\newblock In \emph{International Conference on Learning Representations}.

\bibitem[{Yoon et~al.(2020)Yoon, Zhang, Jordon, and van~der Schaar}]{yoon2020vime}
Yoon, J.; Zhang, Y.; Jordon, J.; and van~der Schaar, M. 2020.
\newblock Vime: Extending the success of self-and semi-supervised learning to tabular domain.
\newblock \emph{Advances in Neural Information Processing Systems}, 33: 11033--11043.

\bibitem[{Zhang(2024)}]{zhang2024gnn}
Zhang, e.~a. 2024.
\newblock Graph Neural Network contextual embedding for Deep Learning on tabular data.
\newblock \emph{Neural Networks (NN)}.

\bibitem[{Zhou et~al.(2022)Zhou, Liu, Chen, Li, Choi, and Hu}]{huang22table2graph}
Zhou, K.; Liu, Z.; Chen, R.; Li, L.; Choi, S.-H.; and Hu, X. 2022.
\newblock Table2Graph: Transforming Tabular Data to Unified Weighted Graph.
\newblock In \emph{Proceedings of the Thirty-First International Joint Conference on Artificial Intelligence, {IJCAI-22}}, 2420--2426.

\bibitem[{Zhu et~al.(2023)Zhu, Shi, Erickson, Li, Karypis, and Shoaran}]{zhu2023xtab}
Zhu, B.; Shi, X.; Erickson, N.; Li, M.; Karypis, G.; and Shoaran, M. 2023.
\newblock XTab: Cross-table Pretraining for Tabular Transformers.
\newblock \emph{arXiv preprint arXiv:2305.06090}.

\end{thebibliography}
